\documentclass{article}

\usepackage{hyperref}       % hyperlinks
\usepackage{url}            % simple URL typesetting
\usepackage{booktabs}       % professional-quality tables
\usepackage{amsfonts}       % blackboard math symbols
\usepackage{nicefrac}       % compact symbols for 1/2, etc.
\usepackage{microtype}      % microtypography
\usepackage{amsmath}
\usepackage{graphicx}
\usepackage{wrapfig}
\usepackage[flushleft]{threeparttable}
\usepackage{multirow}
\DeclareMathAlphabet{\mathpzc}{OT1}{pzc}{m}{it}
\usepackage{amsmath}
\usepackage{mathrsfs}
\usepackage{caption}
\usepackage{subcaption}

\usepackage[accepted]{icml2021}

\icmltitlerunning{TACOS: Task Agnostic Continual Learning in Spiking Neural Networks}

\begin{document}

\twocolumn[
\icmltitle{TACOS: Task Agnostic Continual Learning in Spiking Neural Networks}




\begin{icmlauthorlist}
\icmlauthor{Nicholas Soures}{to}
\icmlauthor{Peter Helfer}{to}
\icmlauthor{Anurag Daram}{to}
\icmlauthor{Tej Pandit}{to}
\icmlauthor{Dhireesha Kudithipudi}{to}
\end{icmlauthorlist}

\icmlaffiliation{to}{Department of Computer Engineering, University of Texas at San Antonio, San Antonio, Texas}

\icmlcorrespondingauthor{Nicholas Soures}{nicholas.soures@utsa.edu}
\icmlcorrespondingauthor{Dhireesha Kudithipudi}{dhireesha.kudithipudi@utsa.edu}

\icmlkeywords{Machine Learning, ICML}

\vskip 0.3in
]



\printAffiliationsAndNotice{}  % leave blank if no need to mention equal contribution

\newcommand{\ph}[1]{{\footnotesize \sf {\color{red}{\bf PH:} [#1]}}}

\begin{abstract}
% This document provides a basic paper template and submission guidelines.
% Abstracts must be a single paragraph, ideally between 4--6 sentences long.
% Gross violations will trigger corrections at the camera-ready phase.
Catastrophic interference, the loss of previously learned information when learning new information, remains a major challenge in machine learning. Since living organisms do not seem to suffer from this problem, researchers have taken inspiration from biology to improve memory retention in artificial intelligence systems. However, previous attempts to use bio-inspired mechanisms have typically resulted in systems that rely on task boundary information during training and/or explicit task identification during inference, information that is not available in real-world scenarios. Here, we show that neuro-inspired mechanisms such as synaptic consolidation and metaplasticity can mitigate catastrophic interference in a spiking neural network, using only synapse-local information, with no need for task awareness, and with a fixed memory size that does not need to be increased when training on new tasks. Our model, TACOS, combines neuromodulation with complex synaptic dynamics to enable new learning while protecting previous information. We evaluate TACOS on sequential image recognition tasks and demonstrate its effectiveness in reducing catastrophic interference. Our results show that TACOS outperforms existing regularization techniques in domain-incremental learning scenarios. We also report the results of an ablation study to elucidate the contribution of each neuro-inspired mechanism separately. 
%Regularization approaches to continual learning can significantly reduce catastrophic interference, but typically rely on task boundary information during training and/or explicit task identification during inference, information that is not available in real-world scenarios. Here, we show that neuro-inspired mechanisms such as synaptic consolidation and metaplasticity can mitigate catastrophic interference in spiking neural networks, using only synapse-local information, without task awareness. Our model, TACOS, combines neuromodulation and complex synaptic dynamics to balance acquisition of new knowledge and protection of old knowledge. We evaluate TACOS on sequential image recognition tasks and demonstrate its effectiveness in reducing catastrophic interference. TACOS enables task-agnostic continual learning, outperforming several existing regularization techniques in domain-incremental learning scenarios, which often grow in memory with the number of tasks or require some form of task awareness.

\end{abstract}

\section{Introduction}
\label{introduction}
When artificial neural networks are presented with new learning tasks, they are prone to forget previously acquired material. This problem, known as \textit{catastrophic interference} (or catastrophic forgetting), was first identified in 1989 \cite{mccloskey_1989_Catastrophic} and continues to be a challenge in the field of artificial intelligence \cite{parisi_2019_Continuala}. By contrast, humans and other mammals do not seem to suffer from this problem. Neuroscientists and machine learning researchers have therefore sought to understand how biological neural systems are able to avoid catastrophic interference, and several explanations have been proposed, %most of them falling into one of the following three categories: (a) rehearsal of previously learned material, e.g. during sleep, prevents memories from being overwritten \cite{mcclelland1995replay, robins_1995_Catastrophic, french_2001_Pseudopatterns}, (b) neurogenesis provides new neural resources for creating memory traces without disturbing existing ones \cite{parisi2018role}, and (c) metaplasticity mechanisms prevent modification of synapses that have been recruited to represent preexisting memories \cite{abraham_1996_Metaplasticity, finnie_2012_role}. 
most of them variations of metaplasticity~\cite{abraham_1996_Metaplasticity, finnie_2012_role}, neurogenesis~\cite{parisi2018role}, or episodic replay~\cite{mcclelland1995replay, robins_1995_Catastrophic, french_2001_Pseudopatterns}. 
Although direct evidence for such hypotheses is difficult to come by, computational modeling has demonstrated that analogous mechanisms can reduce catastrophic interference in artificial neural networks. \cite{meeter_2005_Tracelink, oreilly_2011_Complementary, draelos_2017_Neurogenesis}. 
In this paper we introduce TACOS, a spiking neural network (SNN) that addresses continual learning. Sequential learning of tasks is enabled by error-driven neuromodulation. To mitigate catastrophic forgetting we propose a complex synaptic model that utilizes a form of activity-dependent metaplasticity with synaptic consolidation and heterosynaptic decay. 
Why a spiking model? Spiking neural networks are considered more neurally plausible than rate-based networks~\cite{pfeiffer_2018_Deep}, and are orders of magnitude more energy-efficient~\cite{neftci2017event, nawrocki_2016_Mini}, yet are at least as computationally powerful as rate-based networks \cite{gerstner_2002_Spiking}. Nevertheless, unlike rate-based networks, SNNs have not yet achieved a level of performance suitable for practical applications, mainly because of a lack of efficient and scalable learning algorithms. This situation is, however, beginning to change as new SNN learning algorithms are being introduced, including surrogate-gradient back-propagation \cite{neftci2019surrogate}, three-factor Hebbian learning \cite{fremaux_2016_Neuromodulated}, and homeostatic or non-Hebbian plasticity \cite{watt_2010_Homeostatic}.
%SNNs have also been shown to be more energy-efficient than rate-based networks \cite{neftci2017event, nawrocki_2016_Mini}, and furthermore, spiking network accelerators are being announced with energy advantages over rate-based accelerators \cite{davies_2018_Loihi, debole_2019_TrueNorth}. SNNs have also been shown to be at least as computationally powerful as rate-based networks \cite{gerstner_2002_Spiking}. 
We thus find ourselves at a moment in time when spiking neural networks are coming into their own, and addressing the continual learning problem in the spiking domain is an important step on the way forward. Recent contributions to this effort include investigations of the benefits of local Hebbian learning \cite{munoz2019unsupervised}, Hebbian learning with weight decay \cite{panda2017asp}, and controlled forgetting, a technique that directs plasticity toward the least active regions of the network \cite{allred2020controlled}. These techniques have, however,  been limited to spiking models that are not capable of supervised training, or only able to train a single layer. TACOS does not suffer from these limitations, and also differs from previously published models \cite{ewc,chaudhry2018riemannian, kolouri2020sliced,lee2017overcoming,zenke2017continual} in that \textbf{it does not require any task-identifying information during training or inference, the learning rules are entirely synapse-local, and the algorithm is scalable to multi-layer spiking networks}. %Recently, \citet{rao2019continual} and \citet{aljundi2019task,aljundi2019gradient} proposed rate-based continual learning models that do not require task knowledge during learning, an important step in moving towards designing models that are suitable for general real-world scenarios.

Our results show that TACOS outperforms several state-of-the-art regularization models in the domain-incremental scenario~\cite{vandeven_2019_Three}.
%, and reduces forgetting by 50\% compared to a baseline SNN
 Importantly, TACOS achieves these results 
with a small number of added parameters,
using a fixed memory size that does not depend on the number of tasks to be learned.
%, and without the need for any task-identifying information during training or inference. 

In addition to comparing TACOS to other models, we provide an analysis of TACOS' performance in continual learning by i) studying how the model performs when the amount of training data is reduced, ii) studying how the trade-off between stability and plasticity depends on the degree of metaplasticity, and iii) performing an ablation study to investigate the effects of metaplasticity and consolidation separately.

\section{Background}
%Several research groups have recently proposed solutions to address continual learning, which mainly fall into three categories\cite{parisi_2019_Continuala}: regularization-based approaches (inspired by homeostatic plasticity) which we will talk in-depth about later, memory- and experience-replay techniques (complementary learning systems) \cite{mcclelland1995replay,rolnick2019experience,kemker2017fearnet,lopez2017gradient,aljundi2019gradient}, and architectural approaches such as allocation of new resources for network expansion (inspired by neurogenesis in the hippocampus) \cite{eriksson1998neurogenesis,parisi2018role,sodhani2018training,long2011adaptive,li2019learn,hung2019compacting}, or heterogeneous learning rules \cite{javed2019meta,allred2019stimulating,velez2017diffusion,masse2018alleviating,ellefsen2015neural,marblestone2016toward}. Replay-based techniques require either a secondary generative network or additional memory to store the input representations that allow retention of knowledge of previous tasks. Either approach will also require additional replay phases, during which a system must go offline. On the other hand, neurogenesis-based approaches rely on a potentially indefinite network capacity which is unrealistic in physical systems, though such methods could be paired with some form of pruning to balance overall network size. 

\subsection{Neural Mechanisms}
There is near-universal agreement among neuroscientists that brains store memories in the strengths of synapses~\cite{langille_2018_Synaptic}. The ability of synapses to weaken or strengthen over time is called \textit{synaptic plasticity}, and a number of cellular mechanisms for plasticity have been discovered~\cite{kandel_2014_Molecular}. Among them, long-term potentiation (LTP)~\cite{nicoll_2017_Brief} is the one most studied, and it is considered a major cellular mechanism underlying learning and memory~\cite{frankland_2005_organization,kandel_2014_Molecular}.
%,lisman_2018_Memory,lynch_2004_Longterm, sossin_2008_Molecular,morris_2003_Longterm}. 
%Different phases of LTP have been identified. 
Early-phase LTP (E-LTP) results from moderately strong stimulation and lasts from minutes to hours~\cite{abraham_2003_How,malenka_2004_LTP}; late-phase LTP (L-LTP) requires more intense stimulation and can persist for weeks, months, or even years~\cite{abraham_2002_Induction}. The transition from E-LTP to L-LTP is known as \textit{synaptic consolidation}, and it has been identified with the transformation of short-term memories into long-term memories~\cite{sossin_2008_Molecular,morris_2003_Longterm}. 
%The role of synaptic consolidation in the stabilization of memories is well documented~\cite{nicoll_2017_Brief}.

The increase in memory lifetime (reduced spontaneous decay rate) that characterizes consolidation is accompanied by an increased resistance to memory disruption and modification~\cite{dudai_2004_neurobiology}, corresponding to reduced plasticity at the synaptic level~\cite{richards_2017_Persistence}, a form of \textit{metaplasticity} ("plasticity of synaptic plasticity")~\cite{abraham_1996_Metaplasticity}. Both of these transformations - reduced decay rate and reduced susceptibility to disruption - are important aspects of memory stability.

Another physiological process of importance for learning and memory is \textit{neuromodulation}. Unlike point-to-point synaptic transmission, neuromodulation can simultaneously affect the activity and plasticity of large numbers of neurons through widespread release of neurotransmitters in response to novelty, surprise, reward, etc.~\cite{marder_2012_Neuromodulation}.

%metaplasticity can be \textit{homosynaptic}, affecting only recently activated synapses, or \textit{heterosynaptic}, affecting a neuron's synapses more broadly, not only recently activated ones. An example of heterosynaptic metaplasticity is down-regulation of synaptic plasticity following a period of high postsynaptic activity~\cite{abraham_2008_Metaplasticity}.

%  Our work falls under regularization and metaplasticity approaches to addressing continual learning. 
%There are several examples observed in biological systems where metaplasticity and consolidation can be attributed to retaining knowledge. For example...

\subsection{Related Work}
Many previously published regularization approaches \cite{ewc, zenke2017continual, kolouri2020sliced, chaudhry2018riemannian,ahn2019uncertainty} achieve effects similar to those observed in biology by modifying the loss function to retain information or regularize the model's likelihood distribution to protect important synapses. Techniques like EWC \cite{ewc} and SI \cite{zenke2017continual} select important parameters by using a Fisher Information matrix or tracking synapses' credit in improvement on a task, respectively. The models presented in \cite{chaudhry2018riemannian, kolouri2020sliced,lee2017overcoming} select parameters that preserve the distribution of latent representation of the task. A similar research direction has been to leverage metaplasticity for continual learning. \citet{laborieux2020synaptic} apply a version of metaplasticity to a binary neural network model, making synapses with weights of greater magnitude less plastic, to protect them from modification by subsequent training.
 
%\citet{besold2017neural} and \citet{allred2019stimulating} highlight some potential pitfalls of relying on global mechanisms, such as those used in backpropagation. Recently, \citet{rao2019continual} and \citet{aljundi2019task,aljundi2019gradient} proposed continual learning systems that do not require task knowledge during learning, a necessary step in moving towards continual learning systems applicable to more general real-world scenarios. 

\subsection{Main Contribution}
Using a combination of synaptic consolidation, metaplasticity and neuromodulation, TACOS is able to protect knowledge from catastrophic interference with the use of local information, and without recourse to task awareness, and operates with bounded resources, all of which are important characteristics for systems targeted for real-world deployment. We also demonstrate TACOS effectiveness when constrained to more realistic scenarios where data samples are only seen once. Furthermore, we explore the stability-plasticity trade-off to identify the inflection point where the network is performing at an optimum state and the dependencies of this point on the number and duration of each task.
 
%In contrast with previous work, TACOS provides a solution for protecting synaptic knowledge which does not depend on task awareness, and only requires local information.

\section{Methodology}

\subsection{Problem Formulation}

Continual learning is the ability to learn tasks sequentially, without suffering severe performance loss on previously learned tasks when new ones are learned.
Formally, we define a task, $T^t$, to be a set \{$\mathcal{X}^t,\mathcal{Y}^{t}$\} of ordered pairs of input data points and their corresponding class labels.
The continual learning problem is formulated as maximizing the performance of a system across all tasks, $\mathscr{T}$, when trained sequentially.
\subsection{Preliminaries}
We describe the TACOS framework for continual learning in SNNs in discrete time. The basic building block is a spiking neuron model, the leaky integrate and fire (LIF) neuron, with dynamics described by
%\begin{equation}
%    \label{NeuronEq}
%    \tau_m \frac{d \mathcal{V}_i}{d t} = -\mathcal{V}_i + \mathrm{I}_i\mathrm{R},
%\end{equation}
%\vspace{-1 mm}
%\begin{equation}
%    V[n+1] = V[n] + \frac{dt}{\tau_{mem}}((V_{rest}-V[n])+I[n]R)
%    \label{NeuronEq}
%\end{equation}
\begin{equation}
    V(t+1) = V(t) + \frac{{\Delta}t}{\tau_{mem}}\bigg(\Big(V_{rest}-V(t)\Big)+I(t)R\bigg),
    \label{NeuronEq}
\end{equation}
\noindent where $t$ is the simulation time step counter, ${\Delta}t$ is the length of a time step, the membrane potential $V(t)$ integrates synaptic current $I(t)$ over time, $R$ is the membrane resistance, and $\tau_{mem}$ is the membrane time constant which controls the rates of integration and decay. The synaptic current is a summation of the presynaptic action potentials integrated according to %Equation \ref{CurrentEq3}.
%\begin{equation}
%  \tau_{syn} \frac{d \mathrm{I}_{i}}{d t} = \sum_j \mathpzc{w}_{ij} {\sum\limits_{t_j^f}\delta(t-t_j^f)} - \mathrm{I}_{i},
%\end{equation}
%\begin{equation}
%    I[n + 1] = I[n] + \frac{dt}{\tau_{syn}} (\sum_j  {w}_{j} S_j - I[n])
%    \label{current_estimate}
%\end{equation}
\begin{equation}
    I(t + 1) = I(t) + \frac{{\Delta}t}{\tau_{syn}} \Big(\sum_{j=1}^N  {w}_{j} S_j(t) - I(t)\Big).
    \label{current_estimate}
\end{equation}
%{\sum\limits_{t^f}\delta(t-t^f)}
\noindent 
 At each time step, the synaptic input current, ${I(t)}$, is incremented by the summation of incoming $N$ presynaptic action potentials, $S_j$, which have the value 1 for neurons that have fired and 0 otherwise, weighted by the synaptic strengths, ${w}_{j}$, and decays exponentially towards zero with the synaptic time constant $\tau_{syn}$.
 
\begin{figure*}[!ht] 
\centering
\includegraphics[width=0.8\linewidth]{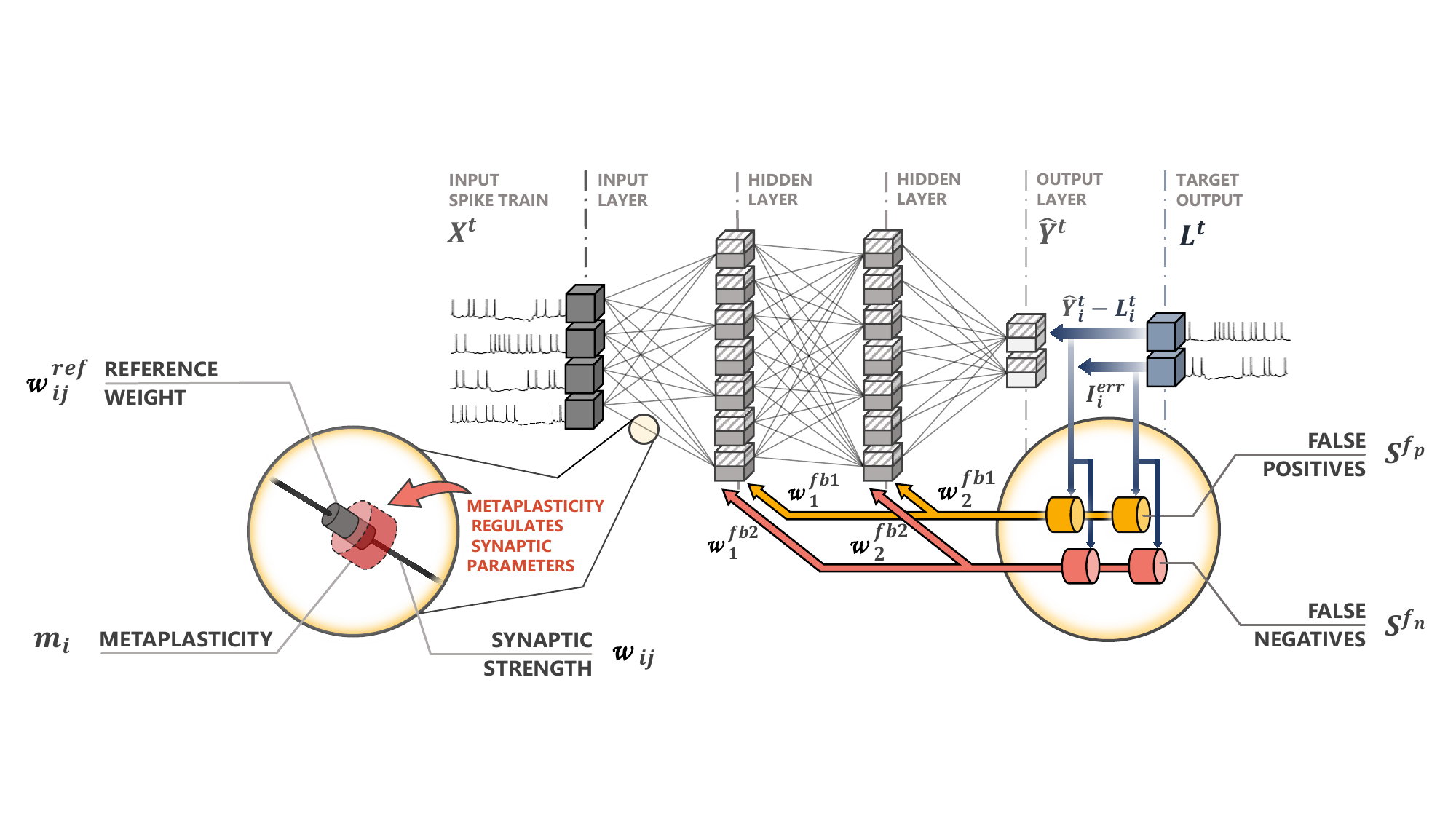}
\caption{Supervised spiking neural network architecture described in this work. Error is propagated directly to all layers through random feedback, where the induction of plasticity is dependent on a Hebbian-like surrogate gradient rule. Unlike traditional neural networks, our model has more complex synapses which consist of the synaptic strength, a reference weight for heterosynaptic plasticity, and activity-dependent metaplasticity which regulates the degree of plasticity per synapse.}
\label{fig:Network}
%\vspace{-5mm}
\end{figure*}

 When the membrane potential crosses a threshold
 $\mathcal{V}_{th}$, an action potential is emitted ($S = 1$) and the membrane potential is reset to zero ($V = 0$). Any neuron that has recently fired undergoes a short time period where its membrane potential is frozen at zero as a simple model of a refractory period. The neural network used in this work is strictly feedforward, as shown in Figure ~\ref{fig:Network}. Input is received as spike trains, one per input neuron, generated from input data, e.g. by Poisson or population encoding.

\subsection{Surrogate Gradient Learning}
To train the model, we implement a surrogate gradient learning rule known as event-driven random backpropagation (eRBP) \cite{neftci2017event}. eRBP relies on presynaptic firing, postsynaptic surrogate gradients, and error feedback through fixed random weights, approximating backpropagation. In eRBP, each neuron has a second compartment $U$ (in addition to the membrane potential $V$)  for integrating the error feedback as described in the following.

At each time step, the difference between the output neurons' spiking activity, ${S}^{out}$, and a target spiking activity or label, ${L}$, is used to generate an error current

\begin{equation}
    {I}^{err}(t) = {S}^{out}(t) - {L}(t).
    \label{Ierr}
\end{equation}

The error current is passed as input to two sets of error-encoding neurons; one set for false positives (neurons that fire when they should not), which receives ${I}^{err}$, and a second set for false negatives (neurons that do not fire when they should), which receives $-{I}^{err}$. The two sets of error-encoding neurons are implemented using integrate-and-fire dynamics, and emit spikes ${S}^{fp}$ and ${S}^{fn}$ when the accumulated error crosses their membrane threshold.

The $i^{th}$ output neuron receives the error signal
\begin{equation}
    E^{o}_i(t) = {S}^{fp}_i(t) - {S}^{fn}_i(t),
\end{equation}
\noindent while the $j^{th}$ hidden neuron in layer $h$ receives the error signal
\begin{equation}
    E^{h}_j(t) = \sum_{i=1}^O \Big({w}^{fp,h}_{i,j}{S}^{fp}_i(t) - {w}^{fn,h}_{i,j}{S}^{fn}_i(t)\Big),
    \label{eq:errfb}
\end{equation}
\noindent where $O$ is the number of output neurons, and ${w}^{fp,h}_{i,j}$ and ${w}^{fn,h}_{i,j}$ are random feedback weights from the error neurons encoding false positives and false negatives, respectively.

The hidden and output neurons integrate the received error signal in a separate compartment $U$, in a manner similar to the way their membrane potentials integrate forward spikes:
\begin{equation}
    U(t+1) = U(t) + \frac{{\Delta}t}{\tau_{mem}}E(t)R.
\end{equation}

Whenever a presynaptic neuron fires, the weights of synapses connecting that neuron to postsynaptic neurons are updated, provided the total synaptic current flowing into the postsynaptic neuron falls in a specified interval, $I_{post}$ $\in$ $\{I_{min}, I_{max}\}$. This condition on post-synaptic current is an approximate surrogate gradient for the LIF neuron based on the relation between LIF firing rates and a ReLU activation, though other surrogate gradient models can be used. The weight update is proportional to the accumulated error in the corresponding post-synaptic neuron's second compartment $U$ (positive or negative):

\begin{equation}
    w_{i,j}(t+1) = w_{i,j}(t) - \eta S_j(t)U_i(t)\Theta(I_i(t)),
\end{equation}

where $S_j$ is a binary value representing if a neuron has fired, $U_i$ is the accumulated error at the post-synaptic neuron of synapse $w_{i,j}$, and $\Theta(I_i)$ is a boxcar function equal to 1 when $I_{post}$ $\in$ $\{I_{min}, I_{max}\}$ and 0 otherwise.

\subsection{Continual Learning with eRBP}

\begin{figure*}[!ht] 
\centering
\includegraphics[width=0.7\linewidth]{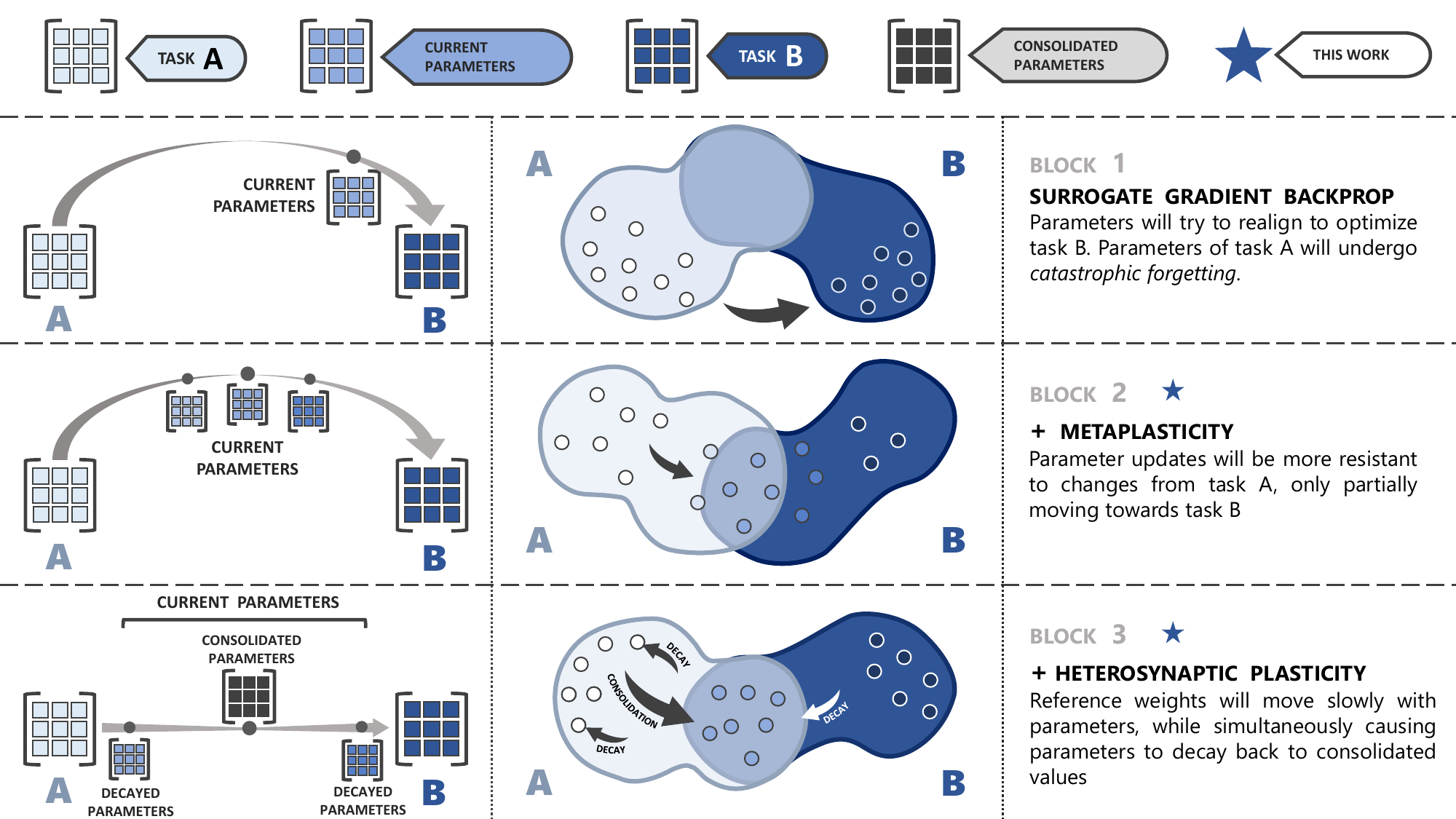}
\caption{ Overview of key plasticity mechanisms for continual learning studied in spiking neural networks. Block 1 illustrates learning in a network without synaptic protection mechanisms. Block 2 (metaplasticity mechanisms) and block 3 (heterosynaptic plasticity) represent two of the new mechanisms incorporated in SNNs for this study.}
\label{fig:Mechanisms}
%\vspace{-5mm}
\end{figure*}

Our objective in the present work is to preserve synaptic knowledge using only information locally available in individual neurons and synapses. To achieve this, we introduce  neuro-inspired mechanisms that preserve synaptic knowledge and add complexity to the synapse itself, shown in Figure \ref{fig:Mechanisms}.

%There are several ways in which this can be addressed, such as bipolar metaplastic updates, or allowing the metaplastic state to decrease or decay. Another approach is to implement neurogenesis when the average metaplastic state in a particular layer crosses a threshold. In this work, we believe that metaplastic state information should ideally be permanent to maintain that knowledge of importance across an indefinite number of tasks, while at the same time it is desirable for the network to not experience unbounded growth in resources. Therefore, to address this concern, we impose a maximum metaplastic state which, although it results in a saturation point where learning is significantly slower on new tasks, still enables synapses to learn continually.

%The advantage of a dynamic metaplastic state is that it allows the network to operate within a range from highly plastic (faster learning) to highly stable (slower learning). This is important because if learning is not fast enough, in lifelong learning scenarios a network will not adapt quickly enough to new information and at the same time highly stable synapse (\textit{i.e.} high metaplastic states) will be hard to change. However, the drawback is that as the network's synapses approach a highly stable state learning will become much slower than using a fixed metaplastic state somewhere in between.

A synapse's plasticity is calculated as a function of its weight and its metaplasticity parameter:

\begin{equation}
    f(m,w) = exp\Big(-abs(mw)\Big),
\end{equation}

\noindent and is used to modulate weight updates, where $w$ is the synaptic weight. %Modelling plasticity as a function of both the metaplastic state and the weight allows low-magnitude synapses to be plastic even when the metaplastic state is high, as this function has a symmetric exponential decrease in plasticity.
\iffalse
\begin{figure}[!ht]
\centering
  \includegraphics[width=0.8\linewidth]{Figures/mw.jpg}
  \caption{Plasticity as a function of the synaptic weight, parameterized by the metaplastic variable $m$.}
\label{fig:mw}
\end{figure}
\fi

The second mechanism is known as \textbf{synaptic consolidation}. The general notion is that synapses have more complexity than can be captured by a single, plastic weight. One approach is to include two weight components that interact with each other and may have different dynamics, e.g. one fast-changing and one slow-changing \cite{zenke2015diverse, leimer_2019_Synaptic, munkhdalai_2020_Sparse}). In TACOS we use a two-component model where in addition to the actual synaptic weight $w$, each synapse has a hidden slower-changing reference weight $w^{ref}$ that tracks the actual weight according to
\begin{equation}\label{weight_decay}
 \mathpzc{w}^{ref}(t+1) = \mathpzc{w}^{ref}(t) + \frac{{\Delta}t}{\tau_{ref}}\Big(w(t) - \mathpzc{w}^{ref}(t)\Big),
\end{equation}
\noindent where $\tau_{ref}$ is the time constant for the evolution of $\mathpzc{w}^{ref}(t)$. The change in $\mathpzc{w}^{ref}(t)$ is driven by its difference from $\mathpzc{w}(t)$.

Whenever the postsynaptic neuron spikes, the actual weights of all its inbound synapses drift towards their reference weights through heterosynaptic plasticity according to
\begin{equation}
    \Delta w_{ij}(t+1) = -\alpha\Big(w_{ij}(t) - \mathpzc{w_{ij}}^{ref}(t)\Big),
\end{equation}
\noindent where $\alpha$ is the decay rate. This decay back towards the reference weight, dependent on the postsynaptic activity, acts as a regularizer for neurons that are actively learning. This helps maintain a balance between the current synaptic changes and consolidated knowledge stored in the reference weight. 

In summary, while metaplasticity slows down synaptic weight changes, it does not prevent long-term shifts in synaptic strengths. Synaptic consolidation works in tandem with metaplasticity to regulate changes in synaptic values: i) slow adjustment of reference weights ensures that repeated plasticity updates, driven by error-modulated plasticity for the current task, will become permanent, ii) small sporadic weight changes, restricted by metaplasticity, are undone by heterosynaptic decay before they have time to consolidate, allowing synaptic weights to remain at values learned from previous tasks. 
%The final weight update in TACOS is:
The complete formula for weight update in TACOS is:

\begin{equation}
\begin{split}
    w_{i,j}(t+1) = w_{i,j}(t) - exp\Big(-abs(m_{i,j}(t)w_{i,j}(t))\Big)\\\bigg(\eta S_j(t)U_i(t)\Theta(I_i(t)) + \alpha\Big(w_{ij}(t) - \mathpzc{w_{ij}}^{ref}(t)\Big)S_i(t)\bigg) ,
\end{split}
\end{equation}

Essentially, the weight update is now a combination of the update from the error-driven surrogate gradient model and a decay term based on the consolidation mechanism. Metaplasticity is used to control the strength of change induced by both error-driven plasticity and decay. These mechanisms come together to mitigate catastrophic interference, relying only on local information.

\section{Results}\label{results}
We evaluate TACOS on split image classification benchmarks using 5-Split MNIST \cite{lecun1998mnist} and Fashion-MNIST \cite{xiao2017fashion} datasets under the task-agnostic domain-incremental continual learning scenario \cite{vandeven_2019_Three, hsu2018re}, where tasks share the same output layer while the \textbf{model is unaware of task identity during both training and inference}. The network configuration was fixed to 200 neurons per hidden layer (1-2 hidden layers), and a two-neuron output layer. \textbf{Unlike most models, TACOS only sees each data sample once (\textit{i.e} one training epoch)}, as would be the case in a real-world scenario~\cite{lopez2017gradient}. 

%Following \cite{lopez2017gradient} and \cite{joseph2020meta}, we use a subset of train samples because; i) a baseline model (metaplasticity, spike-time dependence, and synaptic consolidation were disabled) achieves equivalent performance on traditional classification using either the full or reduced sample size, ii) it is more realistic and challenging to reduce the number of samples, and iii) it reduces simulation time. For simulations, a subset of 8000 training images and the full test set were used.

\textbf{Metrics:} To assess the performance of TACOS and other models, we measure the mean accuracy, $\mathrm{MA}$, across the entire set of tasks $T^1-T^N$, after training on the final task $T^N$: $\mathrm{MA}~=~\sum_{t=1}^{N} \frac{\mathrm{R}^{t,N}}{N}$, where $\mathrm{R}^{t,N}$ is the accuracy on task $T^t$ after training task $T^N$. As a measure of a model's cost we use the memory overhead, $\mathrm{MO}$, calculated as the average amount of memory that a model requires per task, $Mem(T^t)$, in units of the baseline model's memory size $Mem_b$: $\mathrm{MO} = min\Big(1, \frac{1}{N}\sum_{t=1}^{N}\frac{Mem(T^t)}{Mem_b}\Big)$. 

Another important metric when studying catastrophic forgetting is backwards transfer 
%$BWT = \sum_{t=1}^{k} \frac{\mathrm{R}^{t,k}}{k}$, 
$BWT$ = $\frac{1}{k-1}\sum_{t=1}^{k-1}{\mathrm{R}^{t,k}}$, 
the average change in accuracy on tasks $t < k$, after learning task $T^k$. A negative BWT reflects catastrophic interference (the smaller the value, the greater the interference), whereas a positive BWT reflects that training the current task improves the performance on previous tasks. The final metric included in this analysis is forward transfer (FWT) of knowledge; $FWT = \frac{1}{N-k}\sum_{t=k+1}^{N} (\mathrm{R}^{t,k}-b^t)$, the average performance improvement across all tasks $t > k$, after learning task $T^k$, with respect to the baseline performance $b^t$ of the  untrained model. 
%\ph{Maybe $R^{t,0}$ instead of $b^t$?}

\begin{table}[tbh!]
\centering
\begin{threeparttable}
\resizebox{\linewidth}{!}{
\begin{tabular}{@{}c|ccc@{}}
\toprule
            \textbf{Model} & \textbf{FMNIST ($\mathrm{MA}$\%)} & \textbf{MNIST ($\mathrm{MA}$\%)} & \textbf{Memory Overhead ($\mathrm{MO}$)} \\ %\cline{2-7} 
\midrule
Baseline (SNN)      & 75.52 $\pm$ 1.31  &  60.69 $\pm$ 0.6 &       1   \\
SNN + Metaplasticity      & 87.09 $\pm$ 0.96  &  68.59 $\pm$ 7.51 &       1.5x   \\
SNN + Consolidation      & 75.06 $\pm$ 0.88   &  62.10 $\pm$ 0.65 &       2x   \\
TACOS         & \textbf{93.22 $\pm$ 0.22} & 82.56 $\pm$ 1.12   & 2.5x \\   
TACOS - 2 Layer        & 92.94 $\pm$ 0.01 & \textbf{83.45 $\pm$ 0.55}   & 2.5x \\   
%EWC\textendash~\cite{ewc} & 58.85 $\pm$ 2.59 & 63.34 $\pm$ 1.85 & 63.95 $\pm$ 1.90 & 4.57x \\
%SI \textendash~\cite{zenke2017continual}& - & 64.76$\pm$ 3.09 & 65.36 $\pm$ 1.57 & 4x \\
LwF\textendash~\cite{li2017learning}  & 71.02 $\pm$ 0.46& 71.5 $\pm$ 1.63 & 2x \\ 
% MAS\textendash~\cite{MAS} & - & 68.57 $\pm$ 6.85 & 66.42 $\pm$ 2.47  & 3x \\
MAS\textendash~\cite{MAS}  & 68.57 $\pm$ 6.85 & 66.42 $\pm$ 2.47 & 3x \\
Online-EWC\textendash\cite{schwarz2018progress} &  65.55 $\pm$ 3.30 & 64.32 $\pm$ 2.48 & 3x \\
BGD\textendash~\cite{BGD}  & 89.73 $\pm$ 0.88 & 80.44 $\pm$ 0.45 & 3.44x \\
% SS\textendash~\cite{SS} & 49.92 $\pm$ 0.65 & 91.98 $\pm$ 0.12 &  68.23 $\pm$ 2.25 & 2x \\ 
SS\textendash~\cite{SS}  & 91.98 $\pm$ 0.12 &  82.9 $\pm$ 0.01 & 2x \\ 
\bottomrule
\end{tabular}
}
\end{threeparttable}
\caption{Comparison of mean accuracy ($\mathrm{MA}$) and memory overhead ($\mathrm{MO}$) with regularization-based approaches on the Split MNIST and Split Fashion-MNIST(FMNIST) benchmarks. *Each result was averaged across 5 different initializations.}
\label{ExtendedRes}
\end{table}

% Talk about precision of metaplasticity parameter is lower than the weights
% Growth of parameters with respect to the number of tasks is constant for the TACOS models.

\textbf{Continual Learning Analysis:} From Table \ref{ExtendedRes} we can see that TACOS outperforms its similar rate-based counterparts in the Domain-IL scenarios. TACOS is able to achieve this performance with a memory overhead lower than the other models, with the exception of LwF~\cite{li2017learning} and SS~\cite{SS}. It is, however, important to note that the number of parameters or memory overhead of TACOS does not grow with the number of tasks. In order to assess the role of metaplasticity and consolidation, we perform an ablation study by testing the two mechanisms on their own. This can be see in Table \ref{ExtendedRes}, where SNN + Metaplasticity and SNN + Consolidation both perform significantly worse than TACOS.

%Importantly, from the ablation scenarios where only consolidation is used or only metaplasticity, it can be observed that the impact of integrating the two together is significantly better than one or the other.

%As the literature in surrogate gradient models for SNNs continues to advance, models are becoming increasingly capable at benchmarks traditionally limited to rate-based solutions (or models trained in the rate-domain before conversion to spiking). %Future work will aim to extend TACOS to surrogate gradient SNN frameworks demonstrated on CNNs for more complex image datasets such as Cifar and Imagenet. 
%For this work we further study the model by investigating the stability-plasticity trade-offs, impact of reduced data, and impact of task-order on forward and backward transfer using the split-MNIST dataset.

\textbf{TACOS Plasticity-Stability trade-off:} %An important part of addressing catastrophic forgetting is understanding how the model impacts the plasticity-stability trade-off during continual learning. 
To study the plasticity-stability trade-off, we observe learning in three different models. The first model is TACOS, where the maximum metaplastic state is set to different values (see Table~\ref{tab:transfer_table}). The second model is where the metaplastic state is fixed as proposed in \cite{laborieux2020synaptic} with synaptic consolidation (referred to as \textit{Fixed} in Table ~\ref{tab:transfer_table}). The third and last model is the baseline SNN with no metaplasticity or consolidation incorporated.

%We assess this in TACOS with respect to different metaplasticity states, using a constant metaplastic state as proposed in \cite{laborieux2020synaptic} (referred to as \textit{Fixed} in tables and figures), and the baseline SNN.

%%%%%% Previous figure
% \begin{figure*}[!htbp]
% \centering
% \begin{subfigure}{.33\textwidth}
%   \centering
% \includegraphics[width=1\linewidth]{Figures/Task1_Accuracy.png}
% \label{fig:Task1}
% \end{subfigure}%
% \begin{subfigure}{.33\textwidth}
%   \centering
% \includegraphics[width=1\linewidth]{Figures/Task2_Accuracy.png}
% \label{fig:Task2}
% \end{subfigure}
% \begin{subfigure}{.33\textwidth}
%   \centering
% \includegraphics[width=1\linewidth]{Figures/Task3_Accuracy.png}
% \label{fig:Task3}
% \end{subfigure}
% \begin{subfigure}{.33\textwidth}
%   \centering
% \includegraphics[width=1\linewidth]{Figures/Task4_Accuracy.png}
% \label{fig:Task4}
% \end{subfigure}
% \begin{subfigure}{.33\textwidth}
%   \centering
% \includegraphics[width=1\linewidth]{Figures/Task5_Accuracy.png}
% \label{fig:Task5}
% \end{subfigure}
% \caption{Accuracy of each task over time for TACOS with different maximum plasticity settings, a fixed metaplasticity model with different metaplasticity strengths, and a baseline SNN.}
% %\label{fig:test}
% \label{fig:Acc}
% \end{figure*}

\begin{figure*}
\centering
\includegraphics[width=\linewidth]{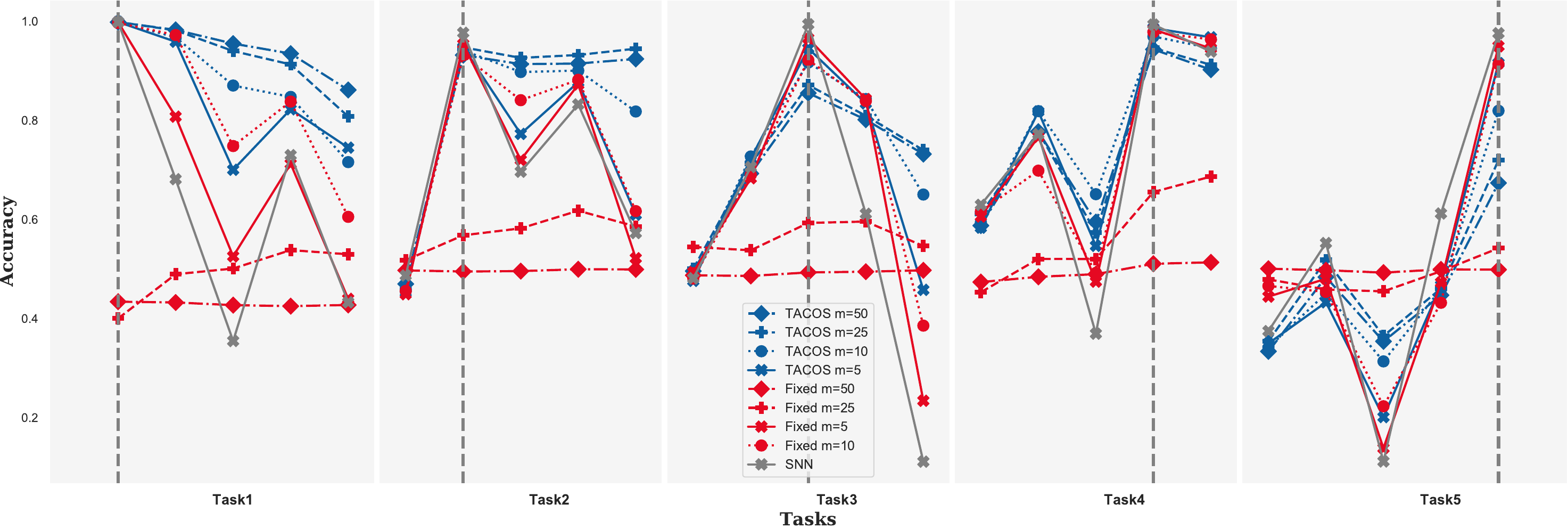}

\caption{Accuracy of each task over time for TACOS with different maximum plasticity settings, a fixed metaplasticity model with different metaplasticity strengths, and a baseline SNN on split-MNIST. The dashed gray line indicates when the model has been trained on the task specified on the x-axis. Accuracy values to the left of the gray line reflect forward transfer induced from learning prior tasks, accuracy values to the right reflect backwards transfer from learning downstream tasks. %\ph{should we also have an "overall" panel, showing mean accuracy of all tasks.}
}
\label{fig:Acc}
\end{figure*}

First, we compare knowledge retention (\textit{i.e.} stability) across models, by measuring the degradation in accuracy on prior tasks. We also monitor the degree of perturbation of the network by measuring the cosine similarity between network representations of the same input after learning each task. 
%Firstly, we compare the retention of knowledge (\textit{i.e.} stability) across models, and assess it in two ways; i) by measuring the degradation in accuracy on prior tasks, and ii) by observing the cosine similarity in network representations for the same input after learning each task. 
As seen in Figure ~\ref{fig:Acc}, TACOS shows lower degradation of accuracy than either the fixed model or the SNN. This result is dramatic when TACOS has a maximum metaplastic state greater than 25; a setting with which the fixed model is incapable of learning. The backwards transfer analysis in Table ~\ref{tab:transfer_table} shows that TACOS has 4-10x less catastrophic interference than the baseline SNN. 
% Can discuss with reference to fixed model

Similarly, the cosine similarity of network representations for the digits in task 1 (\{0,1\}), after learning task 1 and after learning task 5 is 0.98 vs 0.86 on class 1 in TACOS (m=25) compared to the baseline SNN and 0.98 vs 0.85 compared to fixed metaplasticity (m=10) as shown in Table~\ref{tab:order_table}. 

%Though the fixed model with $m=25/50$ shows greater cosine similarity in task 1 representations, this can be misleading because no learning occurs in these models resulting in near random guessing performance unlike TACOS. 

\begin{figure*}[!htbp]
\centering
\begin{subfigure}{.4\textwidth}
  \centering
\includegraphics[width=1\linewidth]{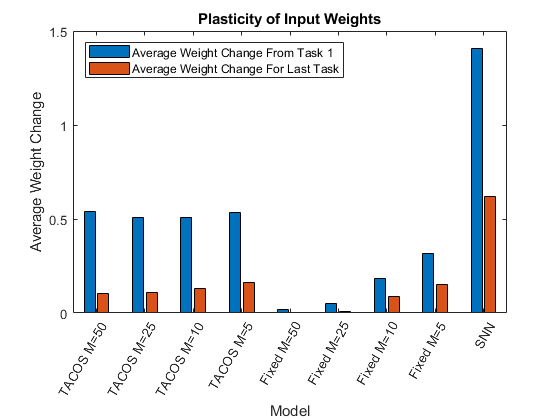}
%\caption{a}
\label{fig:InputWeight}
\end{subfigure}%
\begin{subfigure}{.4\textwidth}
  \centering
\includegraphics[width=1\linewidth]{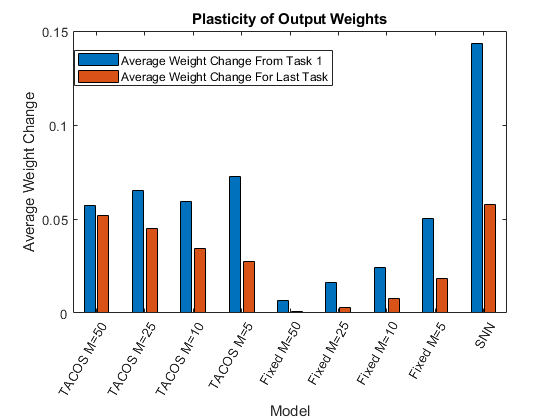}
%\caption{b}
\label{fig:OutputWeight}
\end{subfigure}
\caption{We observe the plasticity in TACOS on split-MNIST by measuring; i) the average weight change induced after learning task 1 (blue), and ii) the average weight change during learning the final task (red). Left) Average change in input-hidden layer weights after learning each task. Right) Average change in hidden-output layer weights after learning each task.}
%\label{fig:test}
\label{fig:Weight}
\end{figure*}

Equally important in addressing catastrophic forgetting is the plasticity of the model, ensuring that the model is capable of learning future tasks. 
To assess the plasticity of each model, we compare the single-task accuracy on the most recently trained task, as well as the average weight change during learning. 
For single-task accuracy (see Table~\ref{tab:transfer_table}), it can be observed that the gap between TACOS and the baseline SNN performance becomes progressively larger as the number of tasks learned increases --- with TACOS performance culminating between $\sim$6\% to $\sim$30\% lower than the baseline SNN on the final task. This can be attributed to metaplasticity and consolidation restricting weight changes to preserve knowledge on prior tasks.
This is also reflected in Figure \ref{fig:Weight}, where the SNN is seen to have $\sim$3x the plasticity of a TACOS model. In comparison to fixed metaplasticity, TACOS allows for significantly more plasticity. Finally, when varying the metaplastic limit in TACOS, there is little change to the average plasticity induced in the network.
%However, among the different maximum metaplastic states, TACOS exhibits the same average amount of plasticity. When comparing TACOS with the fixed metaplastic model, it is apparent that TACOS allows for more plasticity by starting with no metaplasticity. 

%However, an intriguing observation is that for low fixed metaplastic states the single task accuracies are higher than for TACOS with an equivalent maximum metaplastic state in the final task.

%\begin{figure}
%\centering
%\begin{subfigure}{.4\textwidth}
%  \centering
%\includegraphics[width=1\linewidth]{Figures/Class1Similarity.png}
%\label{fig:c1sim}
%\end{subfigure}%
%\begin{subfigure}{.4\textwidth}
%  \centering
%\includegraphics[width=1\linewidth]{Figures/Class2Similarity.png}
%%\caption{b}
%\label{fig:c2sim}
%\end{subfigure}
%\caption{Left) Similarity of average network representation for class 1 (digit 0) over the course of learning on split-MNIST. Right) Similarity of average network representation for class 2 (digit 1) over the course of learning on split-MNIST.}
%%\label{fig:test}
%\label{fig:HiddenSim}
%\end{figure}

\begin{table*}[!htbp]
\resizebox{1\linewidth}{!}{
\begin{tabular}{l|lllllllll}
        & TACOS m=50 & TACOS m=25 & TACOS m=10 & TACOS m=5 & Fixed m=50 & Fixed m=25 & Fixed m=10 & Fixed m=5 & SNN \\ \hline
Class 0 &   0.9641  &   0.9620  &   0.9332   &    0.9188       &    0.9784        &    0.9781    &   0.9756    &   0.9169   & 0.9211 \\
Class 1 &  0.9890   &   0.9872 &   0.9718  &    0.9538       &      0.9775      &    0.9538  &    0.8532 &   0.8240 & 0.8632 \\ \hline
\end{tabular}
}
\caption{Cosine similarity of average hidden layer activity patter for classes 0 and 1 after learning task 1 and task 5. In general, TACOS with m=25 shows the best overall retention in similarity after learning while being negligibly lower than TACOS with m=50 but achieving higher mean accuracy.}
\label{tab:order_table}
\end{table*}

\begin{table*}[!htbp]
\resizebox{\linewidth}{!}{
\begin{tabular}{l|llll|llll|llll|llll|llll}
\toprule
           & \multicolumn{4}{c|}{Task 1} & \multicolumn{4}{c|}{Task 2} & \multicolumn{4}{c|}{Task 3} & \multicolumn{4}{c|}{Task 4} & \multicolumn{4}{c}{Task 5} \\ \cline{2-21} 
           & A & MA      & FWT      & BWT   &  A & MA     & FWT     & BWT     & A & MA     & FWT     & BWT     & A & MA     & FWT     & BWT     &  A & MA     & FWT   & BWT      \\ \midrule
TACOS m=50 & 99.91 &99.91    & -2.64    & 0     & 93.24 &95.79  & 15.32   & -1.56   & 85.57 &90.86   & -1.08   & -3.07   & 94.41 &89.93   & -4.32   & -4.47   & 67.44 &81.94   & 0     & -7.72    \\
TACOS m=25 & 99.94 &99.94    & -1.69    & 0     & 94.83 &96.53   & 17.10   & -1.70   & 87.19 &91.32   & -1.60   & -4.00   & 94.70 &90.08   & -2.98   & -5.44   & 72.05 &82.56   & 0     & -8.98    \\
TACOS m=10 & 99.94 &99.94    & -2.94    & 0     & 94.21 &95.53   & 16.75   & -3.09   & 91.75 &89.56   & -0.27   & -8.60   & 97.01 &89.08   & -3.08   & -8.86   &  82.01 &79.05   & 0     & -17.41   \\
TACOS m=5  &99.94 &99.94    & -3.26    & 0     &94.83 &95.39   & 15.99   & -3.99   &94.59 &80.68   & -11.12  & -23.66  &98.69 &88.08   & -2.34   & -11.91  &91.63 &73.99   & 0     & -27.44   \\
Fixed m=50 &43.47 &43.47    & -0.84    & 0     &49.56 &46.43   & -0.88   & -0.16   &49.38 &47.26   & 0.61    & -0.31   &51.12 &48.31   & 0.89    & 0       &49.94 &48.79   & 0     & 0.12     \\
Fixed m=25 &40.08 &40.08    & 0        & 0     &56.87 &52.94   & 0.75    & 8.94    &59.37 &55.94   & 0.25    & 5.75    &65.61 &60.24   & 0.52    & 6.35    &54.26 &57.88   & 0     & 3.30     \\
Fixed m=10 &99.73 &99.73    & 0.66     & 0     &93.23 &95.27   & 11.61   & -2.43   &92.24 &83.76   & -12.75  & -16.96  &97.89 &88.50   & -5.85   & -9.7    &91.43 &69.77   & 0     & -31.42   \\
Fixed m=5  &99.91 &99.91    & 0        & 0     &95.46 &88.14   & 14.54   & -19.09  &96.62 &73.78   & -17.87  & -35.33  &98.42 &85.35   & -1.77   & -16.34  &95.04 &61.87   & 0     & -44.02   \\
SNN        &99.92 &99.92    & -0.5     & 0     &97.76 &82.98   & 17.88   & -31.73  &99.47 &68.24   & -24.46  & -46.22  &99.41 &79.24   & 12.17   & -26.53  &97.66 &60.69   & 0     & -47.69   \\ \bottomrule
\end{tabular}
}
\caption{Performance of TACOS with a varying maximum metaplastic state $m$, an SNN with a fixed metaplastic state similar to the approach in \cite{laborieux2020synaptic}, and a baseline SNN on the split-MNIST benchmark. Metrics reported are single task accuracy (A), mean accuracy (MA), forward transfer (FWT), and backwards transfer (BWT).}
\label{tab:transfer_table}
\end{table*}

\begin{figure}[!htp] 
\centering
\includegraphics[width=0.8\linewidth]{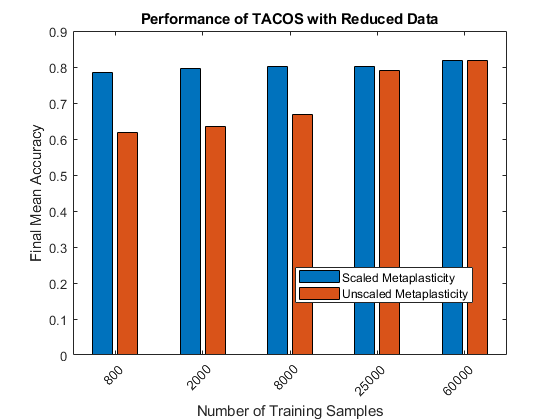}
\caption{Continual learning on split-MNIST with reduced training sets (full test set used for accuracy). Blue accuracy bars represent TACOS with scaled metaplasticity and consolidation based on number of training samples, red accuracy bars reflect using the same parameters that worked best on the full dataset.}
\label{fig:ReducedData}
\vspace{-5mm}
\end{figure}

When considering the optimal trade-off between plasticity and stability there are three main takeaways;

% allows the model greater plasticity than a fixed metaplastic state, and at the same time allows for better retention of knowledge by enabling important synapses to reach high metaplastic states that are not feasible in the fixed metaplasticity model

%the improvement in accuracy for the final task is almost half of that achieved by a baseline SNN. Whereas, catastrophic forgetting is $\sim$5x better in TACOS, while the mean accuracy is XX\% higher than YY.
\begin{itemize}
    \item Using a dynamic metaplastic state as proposed in TACOS shows both greater plasticity and better retention of knowledge than a fixed metaplastic state.
    \item Although the plasticity of TACOS is $\sim$3x lower than that of the baseline SNN, the mean accuracy is $\sim$22\% higher and catastrophic interference is $\sim$5x lower.
    \item The stability-plasticity trade-off in TACOS on the split-MNIST and split-FMNIST tasks shows an inflection point in mean accuracy when the maximum metaplastic state is set to $m=25$.
    
    %is optimal when the maximum metaplastic state is set to $m=25$. Below this point, although plasticity nears the level of the baseline SNN, mean accuracy decreases. Conversely, above this point, even though catastrophic interference is reduced, the mean accuracy also goes down.%Though TACOS with a high metaplastic state of $m=50$ shows slightly better retention of knowledge, the trade-off in plasticity on downstream tasks is much larger than the gains in retention of knowledge compared to a maximum metaplastic state of $m=25$.% It becomes clear that in the split-MNIST task this point is where the plasticity-stability trade-off becomes skewed towards stability, actually resulting in a lower mean accuracy.
\end{itemize}

\begin{figure}[!htp] 
\centering
\includegraphics[width=0.7\linewidth]{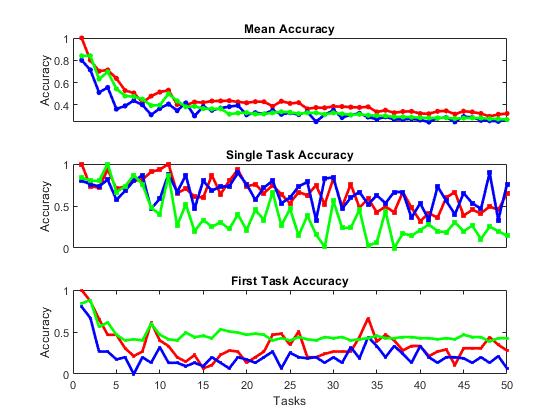}
\caption{Continual learning on the 50-task split-Omniglot dataset. The three scenarios in the plots are: low metaplasticity (red), high metaplasticity (green), and the baseline SNN (blue). The three plots reflect the mean accuracy across all tasks learned \textit{(top)}, the accuracy of the current task \textit{(middle)}, and the accuracy on the first task \textit{(bottom)}.}
\label{fig:Omni}
\vspace{-3mm}
\end{figure}

From this analysis, it is clear that stability increases in TACOS as the number of tasks learned progresses. In the early phases of the split MNIST task the reduction in plasticity is almost negligible, by the final task there is a substantial decrease in the single-task accuracy. In situations where the number of tasks, or even the amount of time spent on each task, is known, the trade-off in plasticity and stability can be optimized by controlling the metaplasticity and consolidation. We demonstrate this by showing the impact on TACOS of reducing the dataset size in Figure \ref{fig:ReducedData}. Some performance degradation is expected due to the reduced number of training samples per task, but the main take-away is that TACOS can largely maintain the same performance, if the metaplasticity and consolidation parameters are scaled based on the time spent learning each task. However, this can become a limitation in deployment on unknown task distributions where potential solutions can be decaying metaplastic updates or scaling metaplastic updates with the magnitude of error on a given sample. Finally, to explore the performance of TACOS over a large family of tasks we create a 50-task Domain-IL scenario with the Omniglot dataset by randomly grouping 5 characters into each task. This is a rather challenging problem for TACOS because each task consists of 75 training images only seen once, leaving very little time for the network to learn the correct patterns, update the metaplastic state, and consolidate information. As seen in Figure \ref{fig:Omni} a low metaplastic state (blue) increases the final mean accuracy by 6\% compared to the baseline SNN (red), while the single task accuracy is relatively similar. Once the metaplasticity is increased by 5x (green), we see that not only does the mean accuracy drop to the same as the baseline SNN, learning of downstream tasks is greatly reduced very early. However, we can see from the first task accuracy that recall of the first task is significantly increased with high metaplasticity, while low metaplasticity barely slows down the forgetting with respect to the baseline SNN.

\section{Conclusion}\label{conclusion}

TACOS results demonstrate that progress towards continual learning in spiking networks is possible using a combination of local plasticity mechanisms. Specifically, we demonstrate that the combination of activity-dependent metaplasticity, synaptic consolidation, heterosynaptic decay, and error-driven neuromodulation can outperform similar rate-based models in domain-IL scenarios. Crucially, TACOS achieves this in a task-agnostic manner, relying only on local information,with a memory budget that does not grow with the number of tasks. By exploring the stability-plasticity trade-off as a function of metaplasticity, we demonstrate that activity-dependent metaplasticity offers significant improvement over a fixed metaplastic state. As learning models in SNNs progress, the mechanisms demonstrated in TACOS will be applicable to spiking CNNs for more complex problems such as Cifar10/100 and ImageNet. In future work, we anticipate to further boost TACOS by  %addressing the limitations in TACOS due to increasing metaplastic states using alternatives to limiting the metaplasticity, such as
: i) allowing for reduction in the metaplastic state, ii) incorporating neurogenesis to introduce neurons with low metaplastic states, or iii) resetting the metaplastic state when long-term consolidation of knowledge can occur through rehearsal.
%Future work will be to address limitations in TACOS due to increasing metaplastic states beyond limiting the metaplasticity such as; i) allowing for reduction in the metaplastic state, ii) neurogenesis, or iii) resetting the metaplastic state when long-term consolidation of knowledge can occur through rehearsal.

%TACOS was able to reduce forgetting in SNNs by 50\%, and compete with state-of-the-art regularization techniques in domain-incremental scenarios. The primary benefits of our approach are minimal overhead in additional parameters compared to other techniques, fixed memory size independent of the number of tasks, and the model is fully task-agnostic.

\vspace{-2mm}
\section*{Acknowledgments}

This material is based on research sponsored by AirForce Research Laboratory under agreement number FA8750-19-S-7010. The U.S. Government is authorized to reproduce and distribute reprints for Governmental purposes notwithstanding any copyright notation thereon. The views and conclusions contained herein are those of the authors and should not be interpreted as necessarily representing the official policies or endorsements, either expressed or implied, of AirForce Research Laboratory or the U.S. Government. 

This work was partially supported through the Lifelong Learning Machines (L2M) program from DARPA/MTO.

\bibliography{references}

\begin{thebibliography}{53}
\providecommand{\natexlab}[1]{#1}
\providecommand{\url}[1]{\texttt{#1}}
\expandafter\ifx\csname urlstyle\endcsname\relax
  \providecommand{\doi}[1]{doi: #1}\else
  \providecommand{\doi}{doi: \begingroup \urlstyle{rm}\Url}\fi

\bibitem[Abraham(2003)]{abraham_2003_How}
Abraham, W.~C.
\newblock How {{Long Will Long}}-{{Term Potentiation Last}}?
\newblock \emph{Philosophical Transactions: Biological Sciences}, 358\penalty0
  (1432):\penalty0 735--744, 2003.
\newblock ISSN 0962-8436.

\bibitem[Abraham \& Bear(1996)Abraham and Bear]{abraham_1996_Metaplasticity}
Abraham, W.~C. and Bear, M.~F.
\newblock Metaplasticity: The plasticity of synaptic plasticity.
\newblock \emph{Trends in Neurosciences}, 19\penalty0 (4):\penalty0 126--130,
  1996.
\newblock ISSN 0166-2236.
\newblock \doi{10.1016/S0166-2236(96)80018-X}.
\newblock URL
  \url{http://www.sciencedirect.com/science/article/pii/S016622369680018X}.

\bibitem[Abraham et~al.(2002)Abraham, Logan, Greenwood, and
  Dragunow]{abraham_2002_Induction}
Abraham, W.~C., Logan, B., Greenwood, J.~M., and Dragunow, M.
\newblock Induction and experience-dependent consolidation of stable long-term
  potentiation lasting months in the hippocampus.
\newblock \emph{Journal of Neuroscience}, 22\penalty0 (21):\penalty0
  9626--9634, November 2002.
\newblock ISSN 0270-6474.

\bibitem[Ahn et~al.(2019)Ahn, Cha, Lee, and Moon]{ahn2019uncertainty}
Ahn, H., Cha, S., Lee, D., and Moon, T.
\newblock Uncertainty-based continual learning with adaptive regularization.
\newblock In \emph{Advances in Neural Information Processing Systems}, pp.\
  4394--4404, 2019.

\bibitem[Aljundi et~al.(2018)Aljundi, Babiloni, Elhoseiny, Rohrbach, and
  Tuytelaars]{MAS}
Aljundi, R., Babiloni, F., Elhoseiny, M., Rohrbach, M., and Tuytelaars, T.
\newblock Memory aware synapses: Learning what (not) to forget.
\newblock In \emph{Proceedings of the European Conference on Computer Vision
  (ECCV)}, pp.\  139--154, 2018.

\bibitem[Allred \& Roy(2020)Allred and Roy]{allred2020controlled}
Allred, J.~M. and Roy, K.
\newblock Controlled forgetting: Targeted stimulation and dopaminergic
  plasticity modulation for unsupervised lifelong learning in spiking neural
  networks.
\newblock \emph{Frontiers in neuroscience}, 14, 2020.

\bibitem[Chaudhry et~al.(2018)Chaudhry, Dokania, Ajanthan, and
  Torr]{chaudhry2018riemannian}
Chaudhry, A., Dokania, P.~K., Ajanthan, T., and Torr, P.~H.
\newblock Riemannian walk for incremental learning: Understanding forgetting
  and intransigence.
\newblock In \emph{Proceedings of the European Conference on Computer Vision
  (ECCV)}, pp.\  532--547, 2018.

\bibitem[Draelos et~al.(2017)Draelos, Miner, Lamb, Cox, Vineyard, Carlson,
  Severa, James, and Aimone]{draelos_2017_Neurogenesis}
Draelos, T.~J., Miner, N.~E., Lamb, C.~C., Cox, J.~A., Vineyard, C.~M.,
  Carlson, K.~D., Severa, W.~M., James, C.~D., and Aimone, J.~B.
\newblock Neurogenesis {{Deep Learning}}.
\newblock \emph{2017 International Joint Conference on Neural Networks
  (IJCNN)}, pp.\  526--533, May 2017.
\newblock \doi{10.1109/IJCNN.2017.7965898}.

\bibitem[Dudai()]{dudai_2004_neurobiology}
Dudai, Y.
\newblock The neurobiology of consolidations, or, how stable is the engram?
\newblock \emph{Annual Review of Psychology}, 55:\penalty0 51--86.
\newblock ISSN 0066-4308.
\newblock \doi{10.1146/annurev.psych.55.090902.142050}.

\bibitem[Finnie \& Nader(2012)Finnie and Nader]{finnie_2012_role}
Finnie, P. S.~B. and Nader, K.
\newblock The role of metaplasticity mechanisms in regulating memory
  destabilization and reconsolidation.
\newblock \emph{Neuroscience and Biobehavioral Reviews}, 36\penalty0
  (7):\penalty0 1667--1707, 2012.
\newblock ISSN 0149-7634.
\newblock \doi{10.1016/j.neubiorev.2012.03.008}.

\bibitem[Frankland \& Bontempi(2005)Frankland and
  Bontempi]{frankland_2005_organization}
Frankland, P.~W. and Bontempi, B.
\newblock The organization of recent and remote memories.
\newblock \emph{Nature Reviews Neuroscience}, 6\penalty0 (2):\penalty0
  119--130, February 2005.
\newblock ISSN 1471-0048.
\newblock \doi{10.1038/nrn1607}.

\bibitem[French et~al.(2001)French, Ans, and
  Rousset]{french_2001_Pseudopatterns}
French, R., Ans, B., and Rousset, S.
\newblock Pseudopatterns and dual-network memory models: {{Advantages}} and
  shortcomings.
\newblock \emph{Connectionist models of learning, development and evolution},
  pp.\  13--22, 2001.

\bibitem[Frémaux \& Gerstner(2016)Frémaux and
  Gerstner]{fremaux_2016_Neuromodulated}
Frémaux, N. and Gerstner, W.
\newblock Neuromodulated {{Spike}}-{{Timing}}-{{Dependent Plasticity}}, and
  {{Theory}} of {{Three}}-{{Factor Learning Rules}}.
\newblock \emph{Frontiers in Neural Circuits}, 9, 2016.
\newblock ISSN 1662-5110.
\newblock \doi{10.3389/fncir.2015.00085}.
\newblock URL
  \url{http://journal.frontiersin.org/Article/10.3389/fncir.2015.00085/abstract}.

\bibitem[Gerstner \& Kistler(2002)Gerstner and Kistler]{gerstner_2002_Spiking}
Gerstner, W. and Kistler, W.~M.
\newblock \emph{Spiking {{Neuron Models}}: {{Single Neurons}}, {{Populations}},
  {{Plasticity}}}.
\newblock {Cambridge University Press}, 2002.
\newblock ISBN 978-0-521-89079-3.
\newblock \doi{10.1017/CBO9780511815706}.
\newblock URL
  \url{https://www.cambridge.org/core/books/spiking-neuron-models/76A3FC77EC2D24CDD91E29EBB23ADB0B}.

\bibitem[Hsu et~al.(2018)Hsu, Liu, Ramasamy, and Kira]{hsu2018re}
Hsu, Y.-C., Liu, Y.-C., Ramasamy, A., and Kira, Z.
\newblock Re-evaluating continual learning scenarios: A categorization and case
  for strong baselines.
\newblock \emph{arXiv preprint arXiv:1810.12488}, 2018.

\bibitem[Kandel et~al.(2014)Kandel, Dudai, and Mayford]{kandel_2014_Molecular}
Kandel, E.~R., Dudai, Y., and Mayford, M.~R.
\newblock The {{Molecular}} and {{Systems Biology}} of {{Memory}}.
\newblock \emph{Cell}, 157\penalty0 (1):\penalty0 163--186, March 2014.
\newblock ISSN 00928674.
\newblock \doi{10.1016/j.cell.2014.03.001}.

\bibitem[Kirkpatrick et~al.(2017)Kirkpatrick, Pascanu, Rabinowitz, Veness,
  Desjardins, Rusu, Milan, Quan, Ramalho, Grabska-Barwinska, et~al.]{ewc}
Kirkpatrick, J., Pascanu, R., Rabinowitz, N., Veness, J., Desjardins, G., Rusu,
  A.~A., Milan, K., Quan, J., Ramalho, T., Grabska-Barwinska, A., et~al.
\newblock Overcoming catastrophic forgetting in neural networks.
\newblock \emph{Proceedings of the national academy of sciences}, 114\penalty0
  (13):\penalty0 3521--3526, 2017.

\bibitem[Kolouri et~al.(2020)Kolouri, Ketz, Pilly, and
  Soltoggio]{kolouri2020sliced}
Kolouri, S., Ketz, N.~A., Pilly, P.~K., and Soltoggio, A.
\newblock Sliced cramer synaptic consolidation for preserving deeply learned
  representations.
\newblock In \emph{International Conference on Learning Representations}, 2020.

\bibitem[Laborieux et~al.(2020)Laborieux, Ernoult, Hirtzlin, and
  Querlioz]{laborieux2020synaptic}
Laborieux, A., Ernoult, M., Hirtzlin, T., and Querlioz, D.
\newblock Synaptic metaplasticity in binarized neural networks.
\newblock \emph{arXiv preprint arXiv:2003.03533}, 2020.

\bibitem[Langille \& Brown(2018)Langille and Brown]{langille_2018_Synaptic}
Langille, J.~J. and Brown, R.~E.
\newblock The {{Synaptic Theory}} of {{Memory}}: {{A Historical Survey}} and
  {{Reconciliation}} of {{Recent Opposition}}.
\newblock \emph{Frontiers in Systems Neuroscience}, 12, 2018.
\newblock ISSN 1662-5137.
\newblock \doi{10.3389/fnsys.2018.00052}.

\bibitem[LeCun et~al.(1998)LeCun, Bottou, Bengio, Haffner,
  et~al.]{lecun1998mnist}
LeCun, Y., Bottou, L., Bengio, Y., Haffner, P., et~al.
\newblock Gradient-based learning applied to document recognition.
\newblock \emph{Proceedings of the IEEE}, 86\penalty0 (11):\penalty0
  2278--2324, November 1998.
\newblock ISSN 0018-9219.
\newblock \doi{10.1109/5.726791}.

\bibitem[Lee et~al.(2017)Lee, Kim, Jun, Ha, and Zhang]{lee2017overcoming}
Lee, S.-W., Kim, J.-H., Jun, J., Ha, J.-W., and Zhang, B.-T.
\newblock Overcoming catastrophic forgetting by incremental moment matching.
\newblock In \emph{Advances in Neural Information Processing Systems}, pp.\
  4652--4662, 2017.

\bibitem[Leimer et~al.(2019)Leimer, Herzog, and Senn]{leimer_2019_Synaptic}
Leimer, P., Herzog, M., and Senn, W.
\newblock Synaptic weight decay with selective consolidation enables fast
  learning without catastrophic forgetting.
\newblock \emph{BioRxiv}, pp.\  613265, 2019.

\bibitem[Li \& Hoiem(2017)Li and Hoiem]{li2017learning}
Li, Z. and Hoiem, D.
\newblock Learning without forgetting.
\newblock \emph{IEEE transactions on pattern analysis and machine
  intelligence}, 40\penalty0 (12):\penalty0 2935--2947, 2017.

\bibitem[Lopez-Paz et~al.(2017)]{lopez2017gradient}
Lopez-Paz, D. et~al.
\newblock Gradient episodic memory for continual learning.
\newblock In \emph{Advances in Neural Information Processing Systems}, pp.\
  6467--6476, 2017.

\bibitem[Malenka \& Bear(2004)Malenka and Bear]{malenka_2004_LTP}
Malenka, R.~C. and Bear, M.~F.
\newblock {{LTP}} and {{LTD}}: {{An}} embarrassment of riches.
\newblock \emph{Neuron}, 44\penalty0 (1):\penalty0 5--21, September 2004.
\newblock ISSN 0896-6273.
\newblock \doi{10.1016/j.neuron.2004.09.012}.

\bibitem[Marder(2012)]{marder_2012_Neuromodulation}
Marder, E.
\newblock Neuromodulation of {{Neuronal Circuits}}: {{Back}} to the {{Future}}.
\newblock \emph{Neuron}, 76\penalty0 (1):\penalty0 1--11, October 2012.
\newblock ISSN 0896-6273.
\newblock \doi{10.1016/j.neuron.2012.09.010}.

\bibitem[McClelland et~al.(1995)McClelland, McNaughton, and
  O'Reilly]{mcclelland1995replay}
McClelland, J.~L., McNaughton, B.~L., and O'Reilly, R.~C.
\newblock Why there are complementary learning systems in the hippocampus and
  neocortex: insights from the successes and failures of connectionist models
  of learning and memory.
\newblock \emph{Psychological review}, 102\penalty0 (3):\penalty0 419, 1995.

\bibitem[McCloskey \& Cohen(1989)McCloskey and
  Cohen]{mccloskey_1989_Catastrophic}
McCloskey, M. and Cohen, N.~J.
\newblock Catastrophic interference in connectionist networks: {{The}}
  sequential learning problem.
\newblock In \emph{Psychology of Learning and Motivation}, volume~24, pp.\
  109--165. {Academic Press}, 1989.
\newblock ISBN 0079-7421.
\newblock URL
  \url{http://www.sciencedirect.com/science/article/pii/S0079742108605368}.

\bibitem[Meeter \& Murre(2005)Meeter and Murre]{meeter_2005_Tracelink}
Meeter, M. and Murre, J. M.~J.
\newblock Tracelink: {{A}} model of consolidation and amnesia.
\newblock \emph{Cognitive Neuropsychology}, 22\penalty0 (5):\penalty0 559--587,
  2005.
\newblock ISSN 0264-3294.
\newblock \doi{10.1080/02643290442000194}.

\bibitem[Morris(2003)]{morris_2003_Longterm}
Morris, R. G.~M.
\newblock Long-term potentiation and memory.
\newblock \emph{Philosophical Transactions of the Royal Society of London
  Series B-Biological Sciences}, 358\penalty0 (1432):\penalty0 643--647, April
  2003.
\newblock ISSN 0962-8436.
\newblock \doi{10.1098/rstb.2002.1230}.

\bibitem[Munkhdalai(2020)]{munkhdalai_2020_Sparse}
Munkhdalai, T.
\newblock Sparse meta networks for sequential adaptation and its application to
  adaptive language modelling.
\newblock \emph{arXiv preprint arXiv:2009.01803}, 2020.

\bibitem[Mu{\~n}oz-Mart{\'\i}n et~al.(2019)Mu{\~n}oz-Mart{\'\i}n, Bianchi,
  Pedretti, Melnic, Ambrogio, and Ielmini]{munoz2019unsupervised}
Mu{\~n}oz-Mart{\'\i}n, I., Bianchi, S., Pedretti, G., Melnic, O., Ambrogio, S.,
  and Ielmini, D.
\newblock Unsupervised learning to overcome catastrophic forgetting in neural
  networks.
\newblock \emph{IEEE Journal on Exploratory Solid-State Computational Devices
  and Circuits}, 5\penalty0 (1):\penalty0 58--66, 2019.

\bibitem[Nawrocki et~al.(2016)Nawrocki, Voyles, and
  Shaheen]{nawrocki_2016_Mini}
Nawrocki, R.~A., Voyles, R.~M., and Shaheen, S.~E.
\newblock A {{Mini Review}} of {{Neuromorphic Architectures}} and
  {{Implementations}}.
\newblock \emph{IEEE Transactions on Electron Devices}, 63\penalty0
  (10):\penalty0 3819--3829, October 2016.
\newblock ISSN 1557-9646.
\newblock \doi{10.1109/TED.2016.2598413}.

\bibitem[Neftci et~al.(2017)Neftci, Augustine, Paul, and
  Detorakis]{neftci2017event}
Neftci, E.~O., Augustine, C., Paul, S., and Detorakis, G.
\newblock Event-driven random back-propagation: Enabling neuromorphic deep
  learning machines.
\newblock \emph{Frontiers in neuroscience}, 11:\penalty0 324, 2017.

\bibitem[Neftci et~al.(2019)Neftci, Mostafa, and Zenke]{neftci2019surrogate}
Neftci, E.~O., Mostafa, H., and Zenke, F.
\newblock Surrogate gradient learning in spiking neural networks: Bringing the
  power of gradient-based optimization to spiking neural networks.
\newblock \emph{IEEE Signal Processing Magazine}, 36\penalty0 (6):\penalty0
  51--63, 2019.

\bibitem[Nicoll(2017)]{nicoll_2017_Brief}
Nicoll, R.~A.
\newblock A {{Brief History}} of {{Long}}-{{Term Potentiation}}.
\newblock \emph{Neuron}, 93\penalty0 (2):\penalty0 281--290, January 2017.
\newblock ISSN 0896-6273.
\newblock \doi{10.1016/j.neuron.2016.12.015}.

\bibitem[O'Reilly et~al.(2011)O'Reilly, Bhattacharyya, Howard, and
  Ketz]{oreilly_2011_Complementary}
O'Reilly, R.~C., Bhattacharyya, R., Howard, M.~D., and Ketz, N.
\newblock Complementary {{Learning Systems}}.
\newblock \emph{Cognitive Science}, pp.\  no--no, December 2011.
\newblock ISSN 03640213.
\newblock \doi{10.1111/j.1551-6709.2011.01214.x}.

\bibitem[Panda et~al.(2017)Panda, Allred, Ramanathan, and Roy]{panda2017asp}
Panda, P., Allred, J.~M., Ramanathan, S., and Roy, K.
\newblock Asp: Learning to forget with adaptive synaptic plasticity in spiking
  neural networks.
\newblock \emph{IEEE Journal on Emerging and Selected Topics in Circuits and
  Systems}, 8\penalty0 (1):\penalty0 51--64, 2017.

\bibitem[Parisi et~al.(2018)Parisi, Ji, and Wermter]{parisi2018role}
Parisi, G.~I., Ji, X., and Wermter, S.
\newblock On the role of neurogenesis in overcoming catastrophic forgetting.
\newblock \emph{arXiv preprint arXiv:1811.02113}, 2018.

\bibitem[Parisi et~al.(2019)Parisi, Kemker, Part, Kanan, and
  Wermter]{parisi_2019_Continuala}
Parisi, G.~I., Kemker, R., Part, J.~L., Kanan, C., and Wermter, S.
\newblock Continual {{Lifelong Learning}} with {{Neural Networks}}: {{A
  Review}}.
\newblock \emph{Neural Networks}, 113:\penalty0 54--71, 2019.
\newblock ISSN 08936080.
\newblock \doi{10.1016/j.neunet.2019.01.012.}
\newblock URL \url{http://arxiv.org/abs/1802.07569}.

\bibitem[Pfeiffer \& Pfeil(2018)Pfeiffer and Pfeil]{pfeiffer_2018_Deep}
Pfeiffer, M. and Pfeil, T.
\newblock Deep {{Learning With Spiking Neurons}}: {{Opportunities}} and
  {{Challenges}}.
\newblock \emph{Frontiers in Neuroscience}, 12:\penalty0 774, October 2018.
\newblock \doi{10.3389/fnins.2018.00774}.

\bibitem[Richards \& Frankland(2017)Richards and
  Frankland]{richards_2017_Persistence}
Richards, B.~A. and Frankland, P.~W.
\newblock The {{Persistence}} and {{Transience}} of {{Memory}}.
\newblock \emph{Neuron}, 94\penalty0 (6):\penalty0 1071--1084, June 2017.
\newblock ISSN 0896-6273.
\newblock \doi{10.1016/j.neuron.2017.04.037}.

\bibitem[Robins(1995)]{robins_1995_Catastrophic}
Robins, A.
\newblock Catastrophic {{Forgetting}}, {{Rehearsal}} and {{Pseudorehearsal}}.
\newblock \emph{Connection Science}, 7\penalty0 (2):\penalty0 123--146, 1995.
\newblock \doi{10.1080/09540099550039318}.

\bibitem[Schug.~Simon(2020)]{SS}
Schug.~Simon, Benzing.~Frederik, S.~A.
\newblock Task-agnostic continual learning via stochastic synapses.
\newblock
  \url{https://sites.google.com/view/cl-icml/accepted-papers?authuser=0}, 2020.
\newblock (Accessed on 09/21/2020).

\bibitem[Schwarz et~al.(2018)Schwarz, Luketina, Czarnecki, Grabska-Barwinska,
  Teh, Pascanu, and Hadsell]{schwarz2018progress}
Schwarz, J., Luketina, J., Czarnecki, W.~M., Grabska-Barwinska, A., Teh, Y.~W.,
  Pascanu, R., and Hadsell, R.
\newblock Progress \& compress: A scalable framework for continual learning.
\newblock \emph{arXiv preprint arXiv:1805.06370}, 2018.

\bibitem[Sossin(2008)]{sossin_2008_Molecular}
Sossin, W.~S.
\newblock Molecular memory traces.
\newblock In Sossin, W.~S., Lacaille, J.~C., Castellucci, V.~F., and
  Belleville, S. (eds.), \emph{Essence of {{Memory}}}, volume 169, pp.\  3--25.
  {Elsevier Science Bv}, {Amsterdam}, 2008.
\newblock ISBN 978-0-444-53164-3.

\bibitem[{van de Ven} \& Tolias(2019){van de Ven} and
  Tolias]{vandeven_2019_Three}
{van de Ven}, G.~M. and Tolias, A.~S.
\newblock Three scenarios for continual learning.
\newblock \emph{arXiv:1904.07734 [cs, stat]}, April 2019.

\bibitem[Watt \& Desai(2010)Watt and Desai]{watt_2010_Homeostatic}
Watt, A.~J. and Desai, N.~S.
\newblock Homeostatic plasticity and stdp: keeping a neuron's cool in a
  fluctuating world.
\newblock \emph{Frontiers in synaptic neuroscience}, 2:\penalty0 5, 2010.

\bibitem[Xiao et~al.(2017)Xiao, Rasul, and Vollgraf]{xiao2017fashion}
Xiao, H., Rasul, K., and Vollgraf, R.
\newblock Fashion-mnist: a novel image dataset for benchmarking machine
  learning algorithms.
\newblock \emph{CoRR}, abs/1708.07747, 2017.
\newblock URL \url{http://arxiv.org/abs/1708.07747}.

\bibitem[Zenke et~al.(2015)Zenke, Agnes, and Gerstner]{zenke2015diverse}
Zenke, F., Agnes, E.~J., and Gerstner, W.
\newblock Diverse synaptic plasticity mechanisms orchestrated to form and
  retrieve memories in spiking neural networks.
\newblock \emph{Nature Communications}, 6\penalty0 (1):\penalty0 1--13, 2015.

\bibitem[Zenke et~al.(2017)Zenke, Poole, and Ganguli]{zenke2017continual}
Zenke, F., Poole, B., and Ganguli, S.
\newblock Continual learning through synaptic intelligence.
\newblock In \emph{Proceedings of the 34th International Conference on Machine
  Learning}, volume~70, pp.\  3987--3995. JMLR. org, 2017.

\bibitem[Zeno et~al.(2018)Zeno, Golan, Hoffer, and Soudry]{BGD}
Zeno, C., Golan, I., Hoffer, E., and Soudry, D.
\newblock Task agnostic continual learning using online variational bayes.
\newblock \emph{arXiv preprint arXiv:1803.10123}, 2018.

\end{thebibliography}
\bibliographystyle{icml2021}

%%%%%%%%%%

\end{document}

\section{Additional Network Details:}
The general dataflow process through our model is shown in Algorithm \ref{SNN_training}. The network applies the continual learning mechanisms to each weight update every time step, while the continual learning parameters are updated at the end of every sample.

\begin{algorithm}
\small
%\DontPrintSemicolon
%\SetKwInOut{Input}{Input}
%\SetKwInOut{Output}{output}
%\SetKw{to}{ to } 
%\SetKw{in}{ in }
%\Input{ Task Distribution$\mathcal{T}$, set of input, target pairs\{$\mathcal{X}^t$,$\mathcal{Y}^{t}$\}}
\begin{algorithmic}

\FOR{$t \in \mathcal{T}$}
\FOR{$epoch=0 \to maxE$}
\FOR{$\{x^t,y^t\}$ \in $\{X^t,Y^t\}$}
\FOR{$\tau \in \bigtau_{sim}$}
\STATE Network Prediction: $\hat{Y}^t = f(X^t)$\; \\
\STATE Random Feedback: $\tau_U \frac{\partial \mathrm{U}}{\partial t} = -\mathrm{U} + E\mathrm{R}_U$\; \\
\STATE Update Neuron Trace: $\frac{d}{dt}X_{tr} = -\frac{X_{tr}}{\tau_{tr}}$ \;  \\
\STATE Update ${w}$ with $\Delta w_{ij} = -f(mw)*\big(eRBP + Decay\big)$\; \\

\ENDFOR
\STATE Update $\mathpzc{w}^{ref}$ with $\frac{{\Delta}t}{\tau_{ref}}\Big(w(t) - \mathpzc{w}^{ref}(t)\Big)$\;
\STATE Update $m$ with $m + {\Delta}m$\;
\ENDFOR
\ENDFOR
\ENDFOR

\end{algorithmic}
\caption{SNN training procedure }
\label{SNN_training}
%\algorithmfootnote{For variable and equation explanations refer to the methodology section in the main document.}
\end{algorithm}

Where $f(mw)$ is the metaplasticity function, $eRBP$ is the weight update induced by the traditional eRBP algorithm and $Decay$ is the weight update induced by heterosynaptic decay. The main parameters used for the SNN in this work are given in Table \ref{tab:param}.
\begin{table*}\centering
%   \begin{center}
  \scriptsize
    \caption{Spiking neural network parameters}
    \label{tab:param}
    \begin{tabular}{l|l|r} % <-- Alignments: 1st column left, 2nd middle and 3rd right, with vertical lines in between
      \toprule
       \textbf{SNN Parameter} &  \textbf{Description} & \textbf{Value}\\
      \midrule
      %\hline
      LR & eRBP learning rate & 1$e^{-2}$\\
      $ V_{th}$ & LIF threshold & 1mV (Hidden) | 2mV (Output) | 2.5mV (Error) \\
      $\tau_{syn}$ & Time-constant for synaptic conductance & 10ms (Input) | 25ms\\
      $\tau_{ref}$ & Time-constant refractory period & 4ms\\
      $\tau_{m}$ & LIF time-constant & 15ms (Hidden) | 25ms (Output) \\
      $\tau_{mu}$ & LIF second compartment time-constant & 15ms\\
      $\tau_{me}$ & Error neuron time-constant & 10ms\\
      $R_m$ & Membrane resistance & 1m$\Omega$ (Hidden) | 5m$\Omega$ (Output)\\
      $R_{mu}$ & Membrane resistance LIF second compartment& 5m$\Omega$\\
      $R_{me}$ & Membrane resistance error neuron & 25m$\Omega$ \\
      $F_{input}$ & Maximum Poisson frequency of input data & 250Hz\\
      $F_{label}$ & Maximum Poisson frequency for target spike trains & 200Hz\\
      $I_{min}$ & Min. synaptic current for weight update & -11\\
      $I_{max}$ & Max. synaptic current for weight update & 13\\
      $\tau_{tr}$ & Time constant for postsynaptic spike trace & 50\\
      $\Delata m$ & Metaplasticity update & 0.04 | 0.004 (output layer)\\
      $m_{th}$ & Threshold of neural trace for metaplasticity update & 6 (Input) | 5 (Hidden) | 2 (Output)\\
      $l_{decay}$ & Strength of heterosynaptic decay & $5e^{-4}$\\
      $t_{cons}$ & Timescale of synaptic consolidation & 25s \\
      \bottomrule
    \end{tabular}
%   \end{center}
\end{table*}

\begin{figure}[!ht]
\centering
  \includegraphics[width=0.8\linewidth]{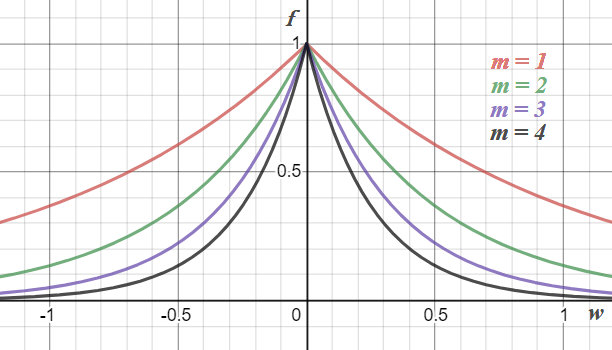}
  \caption{Plasticity as a function of the synaptic weight, parameterized by the metaplastic variable $m$.}
\label{fig:mw}
\end{figure}

\textbf{Metaplasticity Function:} The shape of the metaplasticity function based on the metaplasticity parameter $m$ and weight $w$ is shown in Figure \ref{fig:mw}.

\section{Additional Tests:}
\textbf{Influence of Data Similarity on Continual Learning:} 
An interesting investigation is how aligning tasks consisting of similar/dissimilar information impacts the performance of TACOS. We assess this under five different scenarios:

\begin{itemize}
    \item \textbf{Order 1:} \{0,1\},\{2,3\},\{4,5\},\{6,7\},\{8,9\}: The traditional data split on the split-MNIST continual learning scenario.
    \item \textbf{Order 2:} \{0,1\},\{3,2\},\{5,4\},\{6,7\},\{9,8\}: In the traditional split-MNIST, tasks 2, 3, and 5 had the greatest similarity in classes to task 1 in conflicting output nodes. Therefore, in the second ordering we aim to see if flipping tasks 2, 3, and 5 will show reduced catastrophic interference in task 1.
    \item \textbf{Order 3:} \{4,0\},\{6,8\},\{7,3\},\{9,2\},\{1,5\}: The third order focused on trying to group digits such that in a shared output head, the digits shared among a single output neuron had the greatest similarity.
    \item \textbf{Order 4:} \{0,5\},\{1,7\},\{4,6\},\{8,9\},\{3,2\}: The fourth task was designed to induce catastrophic interference by trying to pair each digit in an opposing category to its most similar digits.
    \item \textbf{Order 5:} \{0,5\},\{1,8\},\{3,2\},\{6,4\},\{9,7\}: The final task was also designed to induce catastrophic interference by trying to place digits in the same group (\textit{i.e.} belonging to the same output neuron), which had the greatest dissimilarity.
\end{itemize}

Using these task orderings, we can observe form Table \ref{tab:order_table} that ordering 2 is not easier than ordering 1 because it only focused on making the problem easier for solving task 1 by switching each pair other than \{6,7\}. The issue with this approach was that in some cases such as task 2 where we highlighted that the greatest similarity was between the digits 0 and 3, the similarity between digits 3 and 1 was also greater than the similarity between 2 and 1. This resulted in order 2 having lower overall performance but it did have much better forward transfer of knowledge than ordering 1 after learning the first task. Ordering 3 showed a 4\% increase in mean accuracy when aligning the data accounting for similarity across all tasks with much more frequent forward transfer of knowledge. Finally, orders 4 and 5 designed to catastrophically interfere showed significantly worse performance in terms of mean accuracy, forward transfer of knowledge, and backwards transfer.

\begin{table*}[]
\resizebox{\linewidth}{!}{
\begin{tabular}{l|lll|lll|lll|lll|lll}
\toprule
        & \multicolumn{3}{c|}{Task 1} & \multicolumn{3}{c|}{Task 2} & \multicolumn{3}{c|}{Task 3} & \multicolumn{3}{c|}{Task 4} & \multicolumn{3}{c}{Task 5} \\ \cline{2-16} 
        & Acc      & FWT      & BWT   & Acc     & FWT     & BWT     & Acc     & FWT     & BWT     & Acc     & FWT     & BWT     & Acc     & FWT   & BWT      \\ \midrule
Order 1 & 99.91    & -2.64    & 0     & 95.79   & 15.32   & -1.56   & 90.86   & -1.08   & -3.07   & 89.93   & -4.32   & -4.47   & 81.94   & 0     & -7.72    \\
Order 2 & 99.92    & 8.69     & 0     & 93.01   & 4.11    & -7.53   & 89.24   & -9.67   & -8.77   & 84.86   & -27.30  & -13.59  & 77.22   & 0     & -13.31   \\
Order 3 & 99.71    & 18.69    & 0     & 96.45   & 3.14    & -3.09   & 93.94   & 8.05    & -3.93   & 91.98   & -16.66  & -4.62   & 86.09   & 0     & -9.74    \\
Order 4 & 98.49    & -12.71   & 0     & 89.27   & -10.37  & -17.38  & 79.76   & -1.52   & -25.11  & 80.97   & -15.21  & -16.99  & 70.48   & 0     & -26.66   \\
Order 5 & 98.61    & -3.51    & 0     & 94.76   & -2.65   & -5.91   & 82.34   & -3.08   & -19.64  & 78.73   & -3.06   & -20.93  & 64.01   & 0     & -34.32   \\ \bottomrule
\end{tabular}
}
\label{tab:order_table}
\caption{Mean accuracy, forwards transfer, and backwards transfer of TACOS $m=50$ with different orderings on the split-MNIST task.}
\end{table*}

\begin{table}[tbh!]
\centering
\begin{threeparttable}
\resizebox{\linewidth}{!}{
\begin{tabular}{@{}c|ccc@{}}
\toprule
            \textbf{Model} & \textbf{FMNIST ($\mathrm{MA}$\%)} & \textbf{MNIST ($\mathrm{MA}$\%)} & \textbf{Memory Overhead ($\mathrm{MO}$)} \\ %\cline{2-7} 
\midrule
Baseline (SNN)      & 75.52 $\pm$ 1.31  &  60.69 $\pm$ 0.6 &       1   \\
SNN + Metaplasticity      & 87.09 $\pm$ 0.96  &  68.59 $\pm$ 7.51 &       1.5x   \\
SNN + Consolidation      & 75.06 $\pm$ 0.88   &  62.10 $\pm$ 0.65 &       2x   \\
TACOS (HF)         & 93.22 $\pm$ 0.22 & 82.56 $\pm$ 1.12   & 2.5x \\   
TACOS (HF) - 2 Layer        & 92.94 $\pm$ 0.01 & 83.45 $\pm$ 0.55   & 2.5x \\  
TACOS (LF)         & 90.05 $\pm$ 0.53 & 81.84 $\pm$ 0.81  & 2.5x \\   
TACOS (LD)    & \textbf{92.48 $\pm$ 0.10} & 82.15 $\pm$ 0.73   & 2.5x \\  
NACHOS (LD)    & -- & \textbf{83.63 $\pm$ 0.1}   & 2.5x \\  
%EWC\textendash~\cite{ewc} & 58.85 $\pm$ 2.59 & 63.34 $\pm$ 1.85 & 63.95 $\pm$ 1.90 & 4.57x \\
%SI \textendash~\cite{zenke2017continual}& - & 64.76$\pm$ 3.09 & 65.36 $\pm$ 1.57 & 4x \\
LwF  & 71.02 $\pm$ 0.46& 71.5 $\pm$ 1.63 & 2x \\ 
% MAS\textendash~\cite{MAS} & - & 68.57 $\pm$ 6.85 & 66.42 $\pm$ 2.47  & 3x \\
MAS  & 68.57 $\pm$ 6.85 & 66.42 $\pm$ 2.47 & 3x \\
Online-EWC &  65.55 $\pm$ 3.30 & 64.32 $\pm$ 2.48 & 3x \\
BGD & 89.73 $\pm$ 0.88 & 80.44 $\pm$ 0.45 & 3.44x \\
% SS\textendash~\cite{SS} & 49.92 $\pm$ 0.65 & 91.98 $\pm$ 0.12 &  68.23 $\pm$ 2.25 & 2x \\ 
SS  & 91.98 $\pm$ 0.12 &  82.9 $\pm$ 0.01 & 2x \\ 
\bottomrule
\end{tabular}
}
\end{threeparttable}
\caption{Comparison of mean accuracy ($\mathrm{MA}$) and memory overhead ($\mathrm{MO}$) with regularization-based approaches on the Split MNIST and Split Fashion-MNIST(FMNIST) benchmarks. For TACOS results H/L represents high-firing rate solutions where neurons may be instantly firing at maximum rates, F/D represents fixed complex synapses vs dynamic synapses presented in this work.}
\label{ExtendedRes}
\end{table}

% In \cite{lopez2017gradient} and \cite{joseph2020meta}, a subset of train samples is used to be more realistic and challenging. T